\documentclass[10pt,twocolumn,letterpaper]{article}

\usepackage{cvpr}
\usepackage{times}
\usepackage{epsfig}
\usepackage{graphicx}
\usepackage{amsmath}
\usepackage{amssymb}

\usepackage[table]{xcolor}
\usepackage{url}
\usepackage{algorithm}
\usepackage[noend]{algpseudocode}
\usepackage{epstopdf}
\usepackage{subfigure}
\usepackage{xspace}
\usepackage{booktabs}
\usepackage{multirow}
\usepackage{siunitx}

\usepackage[pagebackref=true,breaklinks=true,letterpaper=true,colorlinks,bookmarks=false]{hyperref}

\cvprfinalcopy 

\def\cvprPaperID{3202} 

\ifcvprfinal\fi
\begin{document}

\title{Improving Facial Attribute Prediction using Semantic Segmentation}

\author{Mahdi M. Kalayeh\\
	{\tt\small Mahdi@eecs.ucf.edu}
	\and
	Boqing Gong\\
	{\tt\small bgong@crcv.ucf.edu}
	\and
	Mubarak Shah\\
	{\tt\small shah@crcv.ucf.edu}
	\and
	Center for Research in Computer Vision\\
	University of Central Florida\\
}

\maketitle

\begin{abstract}
Attributes are semantically meaningful characteristics whose applicability widely crosses category boundaries. They are particularly important in describing and recognizing concepts where no explicit training example is given, \textit{e.g., zero-shot learning}. Additionally, since attributes are human describable, they can be used for efficient human-computer interaction. In this paper, we propose to employ semantic segmentation to improve facial attribute prediction. The core idea lies in the fact that many facial attributes describe local properties. In other words, the probability of an attribute to appear in a face image is far from being uniform in the spatial domain. We build our facial attribute prediction model jointly with a deep semantic segmentation network. This harnesses the localization cues learned by the semantic segmentation to guide the attention of the attribute prediction to the regions where different attributes naturally show up. As a result of this approach, in addition to recognition, we are able to localize the attributes, despite merely having access to image level labels (weak supervision) during training. We evaluate our proposed method on CelebA  and LFWA datasets and achieve superior results to the prior arts. Furthermore, we show that in the reverse problem, semantic face parsing improves when facial attributes are available. That reaffirms the need to jointly model these two interconnected tasks.
\end{abstract}

\vspace{-4mm}
\section{Introduction}
Nowadays, state-of-the-art computer vision techniques allow us to teach machines different classes of objects, actions, scenes, and even fine-grained categories. However, to learn a certain notion, we usually need positive and negative examples from the concept of interest. This creates a set of challenges as the examples of different concepts are not equally easy to collect. Also, the number of learnable concepts is linearly capped by the cardinality of the training data. Therefore, being able to robustly learn a set of \textit{sharable concepts} that go beyond rigid category boundaries is of tremendous importance. Visual attributes are one particular type of the \textit{sharable concepts}. They are human describable and machine detectable. The fact that attributes are generally not category-specific suggests that one can potentially describe an  exponential number of categories with various combinations of attributes. Naturally, attributes are ``additive'' to the objects (\textit{e.g.}, horn for cow). It means that an instance of an object may or may not take a certain attribute while in either case the category label is preserved (\textit{e.g.}, a cow with or without horn is still a cow). Hence, attributes are especially useful in problems that aim at modeling intra-category variations such as fine-grained classification. 

Despite their additive character, attributes do not appear in arbitrary regions of the objects (\textit{e.g.}, the horn, if appears, would show up on a cow's head). This notion is the basis of our work. That is, in order to detect an attribute, instead of the entire spatial domain, we should focus on the region in which that attribute naturally shows up. We hypothesize that the attribute prediction can benefit from localization cues. However, attribute prediction benchmarks come with holistic image level labels. In addition, sometimes it is hard to define a spatial boundary for a given attribute. For instance, it is not clear that according to which spatial region in a face one decides if a person is ``attractive" or not. To tackle this challenge, we transfer localization cues from a relevant auxiliary task to the attribute prediction problem. 

Using bounding box to show the boundary limits of an object is a common practice in computer vision. However, regions that different attributes occupy drastically change in shape and form. For example, in a face image, one cannot effectively put a bounding box around the region associated to ``hair". In fact, the shape of the region can be used as an indicative signal on the attribute. Therefore, we need an auxiliary task that learns detailed localization information without restricting the corresponding regions to be in certain pre-defined shapes. 

Semantic segmentation has all the aforementioned characteristics. It is the problem of assigning class labels to every pixel in an image. As a result, a successful semantic segmentation approach has to learn pixel-level localization cues which implicitly encode color, structure, and geometric characteristics in fine detail. In this work, we are interested in facial attributes. Hence, the semantic face parsing problem \cite{smith2013exemplar} is a suitable candidate to serve as an auxiliary task to spatially hint the attribute prediction methods.

To perform attribute prediction, we feed an image to a fully convolutional neural network which generates feature maps that are ready to be aggregated \cite{lin2013network} and passed to the classifier. However, global pooling \cite{lin2013network} is agnostic to where, in spatial domain, the attribute-discriminative activations occur. Hence, instead of propagating the attribute signal to the entire spatial domain, we funnel them into the semantic regions. By doing so, our model learns \textit{where} to attend and \textit{how} to aggregate the feature map activations. We refer to this approach as Semantic Segmentation-based Pooling (SSP) where activations at the end of the attribute prediction pipeline are pooled within different semantic regions.

Alternatively, we can incorporate the semantic segmentation into earlier layers of the attribute prediction network with a gating mechanism. Specifically, we augment the max pooling operation such that it does not mix activations that reside in different semantic regions. To do so, we gate the activation output of the last convolution layer prior to the max pooling by element-wise multiplying it with the semantic regions. This generates multiple versions of the activation maps that are masked differently and presumably discriminative for various attributes.  We refer to this approach as Semantic Segmentation-based Gating (SSG).

Since the semantic segmentation is not available for the attribute benchmarks, we learn to \textit{estimate} it using a deep semantic segmentation network. Our approach is conceptually similar to \cite{noh2015learning} in which an encoder-decoder model is built using convolution and deconvolution layers. However, considering the relatively small number of available data for the auxiliary segmentation problem, we modify the network architecture in order to adapt it to our facial attribute prediction problem. Despite being much simpler than \cite{noh2015learning}, we found our semantic segmentation network to be very effective in solving the auxiliary task of semantic face parsing. Once trained, such network is able to provide localization cues in the form of semantic segmentation (decoder output) that decompose the spatial domain of an image into mutually exclusive semantic regions.

We show that both SSP and SSG mechanisms outperform the existing state-of-the-art facial attribute prediction techniques while employing them together results in further improvements.

\section{Related Work}\label{sec:relatedwork}
It is fair to say that the attribute prediction literature can be divided into holistic and part-based approaches. The common theme among the holistic methods is to take the entire image into account when extracting features for attribute prediction. On the other hand, part-based methods begin with an attribute-related part detection and then use the localized parts, in isolation from the rest of the image, to extract features.

Our proposed method falls between the two ends of the spectrum. While we process the image in a holistic fashion to generate feature vectors for the classifiers, we employ localization cues in the form of semantic segmentation.

It has been shown that part-based models generally outperform the holistic methods. However, they are prone to the localization error as it can affect the quality of extracted features. Among earlier works we refer to \cite{kumar2009attribute, berg2013poof, bourdev2011describing} as successful examples of part-based attribute prediction approaches. More recently, in an effort to combine part-based models with deep learning, Zhang \textit{et al.} \cite{zhang2014panda} proposed PANDA, a pose-normalized convolutional neural network (CNN) to infer human attributes from images. PANDA employs poselets \cite{bourdev2011describing} to localize body parts and then extracts CNN features from the localized regions. These features will later be used to train SVM classifiers for attribute prediction. Inspired by \cite{zhang2014panda} while seeking to also leverage the holistic cues, Gkioxari \textit{et al.} \cite{gkioxari2015actions} proposed a unified framework that benefits from both holistic and part-based clues while utilizing a deep version of poselets \cite{bourdev2011describing} as part detectors. Liu \textit{et al.} \cite{liu2015faceattributes} have taken a relatively different approach. They show that pre-training on massive number of object categories and then fine-tuning on image level attributes is sufficiently effective in localizing the entire face region. Such weakly supervised method provides them with a located region where they perform facial attribute prediction. Finally, in a part-based approach, Singh \textit{et al.} \cite{singh2016end} use spatial transformer networks \cite{jaderberg2015spatial} to locate the most relevant region associated to a given attribute. They encode such localization cue in a Siamese architecture to perform localization and ranking for relative attributes. 

\section{Methodology}\label{sec:methodology}
In this section, we begin with the attribute prediction models assuming that the semantic regions are given. We then move on to the semantic segmentation network and provide details on how the semantic regions are generated.

\subsection{Attribute Prediction Networks}
To leverage the localization cues for facial attribute prediction, we propose semantic segmentation-based pooling and gating mechanisms. We describe our basic attribute prediction model. Then, we explain SSP and SSG in detail including how they are employed in the basic model, simply as new layers, to improve facial attribute prediction.

\subsubsection{Basic Attribute Prediction Network}
Our basic attribute prediction model is a 12-layers deep fully convolutional neural network. We gradually increase the number of convolution filters from 64 to 1024 filters as we proceed towards the deeper layers. Prior to any increase in the number of convolution filters, we reduce the size of the activation maps using max pooling. For such operation both the kernel size and stride values are set to 2. In our architecture, every convolution layer is followed by the Batch Normalization \cite{ioffe2015batch} and PReLU \cite{he2015delving}. The kernel size and stride values of all the convolution layers are respectively set to 3 and 1. The first 8 layers of our basic attribute prediction network are similar in configuration to the encoder part of the semantic segmentation network and detailed in Table \ref{tab:segnetconfig}. The rest consists of 4 convolution layers of 512 and 1024 filters, two layers of each. At the end of the pipeline, we aggregate the activations of the last convolution layer using global average pooling \cite{lin2013network} to generate 1024-D vector representations. These vectors are subsequently passed to the classifier for attribute prediction. We train the network using sigmoid cross entropy loss. Section \ref{sec:implementationdetails} provides further details on the training procedure.

\subsubsection{SSP: Semantic Segmentation-based Pooling}
We argue that attributes usually have a natural correspondence to certain regions within the object boundary. Hence, aggregating the visual information from the entire spatial domain of an image would not capture this property. This is the case for the global average pooling \cite{lin2013network} used above in our basic attribute prediction model as it is agnostic to where, in the spatial domain, activations occur. Instead of pooling from the entire activation map, we propose to first decompose the activations of the last convolution layer into different semantic regions and then aggregate only those that reside in the same region. Hence, rather than a single 1024-D vector representation, we obtain multiple features, each representing only a single semantic region. This approach has an interesting intuition behind it. In fact, SSP funnels the backpropagation of the label signals, via multiple paths, associated with different semantic regions, through the entire network. This is in contrast with global average pooling that rather equally affects different locations in the spatial domain. We later explore this by visualizing the activation maps of the final convolution layer. 

\begin{figure}
	\centering
	\includegraphics[width=0.47\textwidth]{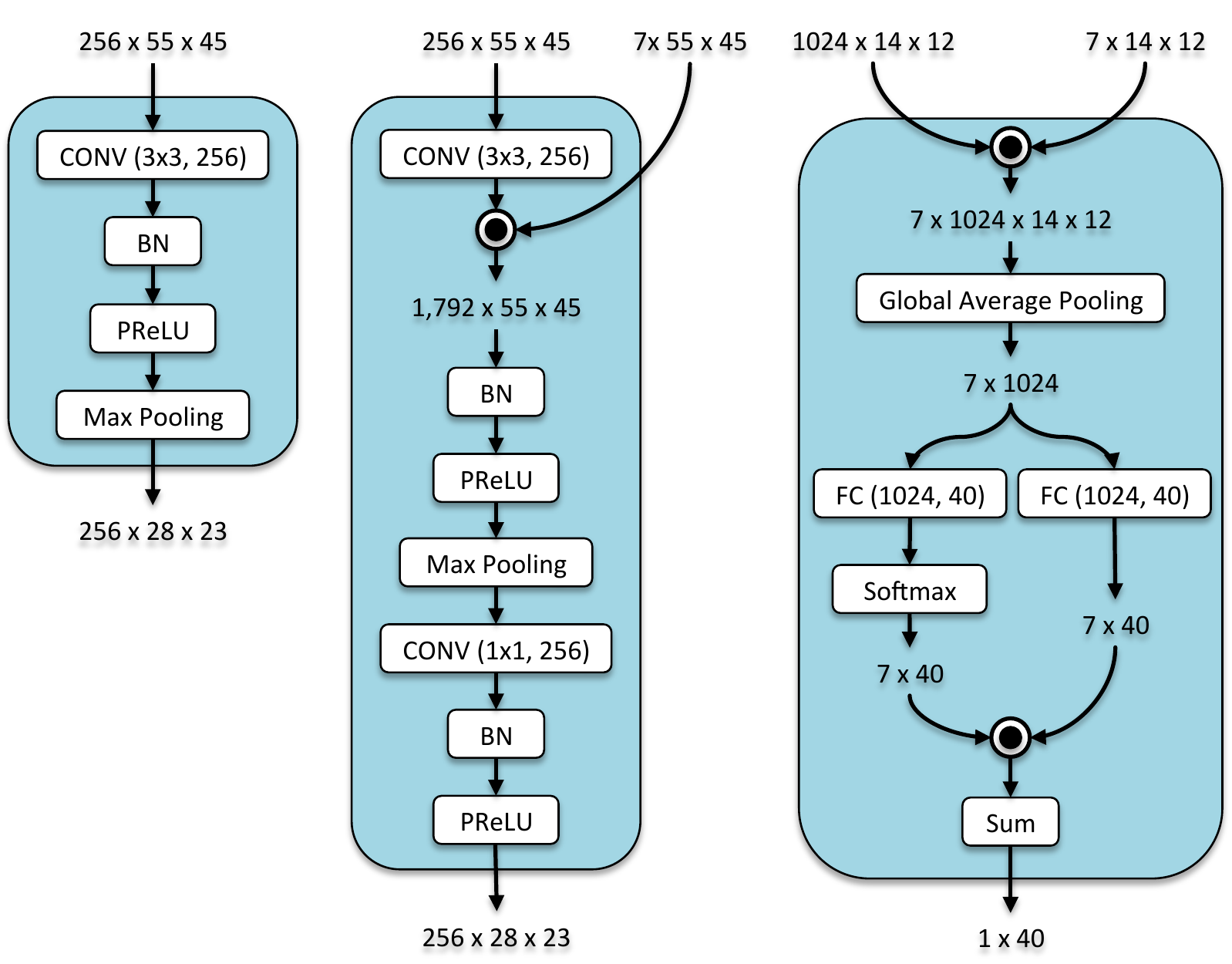}
	\caption{Left: Standard convolution layer followed by max pooling, Middle: SSG, Right: SSP. Note: In this work, there are 7 semantic regions and 40 attributes to predict.}\label{fig:architecures}
	\vspace{-4mm}
\end{figure}

While we can simply concatenate the representations associated with different regions and pass it to the classifier, it is interesting to observe if attributes indeed prefer one semantic region to another. Also, whether what our model learns matches human expectation on what attribute corresponds to which region. To do so, we take a similar approach to \cite{Bilen16} where Bilen and Vedaldi employed a two branch network for weakly supervised object detection. We pass the vector representations, each associated to a different semantic region, to two branches one for recognition and another for localization. We implement these branches as linear classifiers that map 1024-D vectors to the number of attributes. Hence, we have multiple detection scores for an attribute each inferred based on one and only one semantic region. To combine these detection scores, we begin by normalizing the output of the localization branch using softmax non-linearity across different semantic regions. This is a per-attribute operation, not an across-attribute one. We then compute the final attribute detection score by a weighted sum of the recognition branch outputs using weights generated by the localization branch. Figure \ref{fig:architecures}, on the right, shows the SSP architecture.  

\subsubsection{SSG: Semantic Segmentation-based Gating}
The max pooling is used to compress the visual information in the activation maps of the convolution layers. Its efficacy has been proven in many computer vision tasks such as image classification and object detection. However, attribute prediction is inherently different from image classification. In image classification, we want to aggregate the visual information across the entire spatial domain to come up with a single label for the image. Unlike that, many attributes are inherently localized to image regions. Consequently, aggregating activations that reside in the ``hair" region with the ones that correspond to ``mouth'', would confuse the model in detecting ``smiling" and ``wavy hair" attributes. We propose SSG to cope with this challenge. 

Figure \ref{fig:architecures} shows a standard convolution layer followed by max pooling on the left, and the SSG architecture in the middle. The latter is our proposed alternative to the former. Here we assume the convolution layer to preserve the number of input channels but it does not have to be. To gate the output activations of the convolution layer, we broadcast element-wise multiplication for each of the $N=7$ semantic regions with the entire activation maps. This generates $N$ copies (totally $1,792 = 256\times7$ activation maps) of the activations that are masked differently. Such mechanism spatially decomposes the activation maps into copies where activations with high values cannot simultaneously occur in two semantically different regions. For example, gating with the semantic segmentation that corresponds to the mouth region, would suppress the activations falling outside its area while preserving those that reside inside it. However, the area which a semantic region occupies varies from one image to another. 

We observed that, directly applying the output of the semantic segmentation network results in instabilities in the middle of the network. To alleviate this, prior to the gating procedure, we normalize the semantic masks such that the values of each channel sum up to 1. We then gate the activations right after the convolution and before the Batch Normalization \cite{ioffe2015batch}. This is very important since the Batch Normalization \cite{ioffe2015batch} enforces a normal distribution on the output of the gating procedure. Then, we can apply max pooling on these gated activation maps. Since, given a channel, activations can only occur within a single semantic region, max pooling operation cannot blend activation values that reside in different semantic regions. We later restore the number of channels using a 1$\times$1 convolution. It is worth noting that SSG can mimic the standard max pooling by learning  a sparse set of weights for the 1$\times$1 convolution. In a nutshell, semantic segmentation-based gating allows us to process the activations of convolution layers in a per-semantic region fashion, and directly learns how to combine the pooled values afterwards.

\subsection{Semantic Segmentation Network}\label{sec:methodology_segnet}
We have previously explained the rationale behind employing semantic face parsing to improve facial attribute prediction. Our design for the semantic segmentation network follows an encoder-decoder approach, similar in concept to the deconvolution network proposed in \cite{noh2015learning}. However, considering the limited number of training data for the segmentation network, we have made different design decisions to reduce the complexity of the model while preserving its capabilities. The encoder consists of 8 convolution layers in blocks of 2, separated with 3 max pooling layers. This is much smaller than the 13 layers used in the deconvolution network \cite{noh2015learning}. At the end of the encoder part, rather than collapsing the spatial resolution as in \cite{noh2015learning}, we maintain it at the scale of one-eighth of the input size. The decoder is a mirrored version of the encoder replacing convolution layers with deconvolution and max pooling layers with upsampling. Unlike \cite{noh2015learning} that uses switch variables to store the max pooling locations, we simply upsample the activation maps (repetition with  nearest neighbor interpolation). We increase (decrease) the number of convolution (deconvolution) filters by a factor of 2 after each max pooling (upsampling), starting from 64 (512) filters as we proceed along the encoder (decoder) path. Every convolution and deconvolution layer is followed by Batch Normalization \cite{ioffe2015batch} and PReLU \cite{he2015delving}. To cope with the challenge of relatively small number of training data, we propagate the semantic segmentation loss at different depths along the decoder path. That is, before each upsampling layer, we compute the loss by predicting the semantic segmentation maps at different scales. We then aggregate these losses with equal weights prior to backpropagation.  Finally, while \cite{noh2015learning} employs VGG16 \cite{simonyan2014very} weights to initialize the encoder, we train our network from scratch. These design decisions allow us to successfully train the semantic segmentation network with the limited number of training data. Detailed configuration of the semantic segmentation network is shown in Table \ref{tab:segnetconfig}.

\begin{table}\tiny
	\centering
	\resizebox{\columnwidth}{!}{
		\begin{tabular}{|l|l|l|}
			\hline
			\textbf{Layer} & \textbf{Operations} & \textbf{Output size}\\	
			\hline\hline
			Conv$_{11}$ & Conv, BN, PReLU & 64$\times$218$\times$178\\
			Conv$_{12}$ & Conv, BN, PReLU & 64$\times$218$\times$178\\
			MaxPool$_{1}$ & Max Pooling & 64$\times$109$\times$89\\	
			\hline
			Conv$_{21}$ & Conv, BN, PReLU & 128$\times$109$\times$89\\
			Conv$_{22}$ & Conv, BN, PReLU & 128$\times$109$\times$89\\
			MaxPool$_{2}$ & Max Pooling & 128$\times$55$\times$45\\
			\hline
			Conv$_{31}$ & Conv, BN, PReLU & 256$\times$55$\times$45\\
			Conv$_{32}$ & Conv, BN, PReLU & 256$\times$55$\times$45\\
			MaxPool$_{3}$ & Max Pooling & 256$\times$28$\times$23\\
			\hline		
			Conv$_{41}$ & Conv, BN, PReLU & 512$\times$28$\times$23\\
			Conv$_{42}$ & Conv, BN, PReLU & 512$\times$28$\times$23\\
			\hline\hline
			Deconv$_{41}$ & Deconv, BN, PReLU & 512$\times$28$\times$23\\
			Deconv$_{42}$ & Deconv, BN, PReLU & 512$\times$28$\times$23\\
			\hline
			UpSample$_{3}$ & UpSampling & 512$\times$55$\times$45\\
			Deconv$_{31}$ & Deconv, BN, PReLU & 256$\times$55$\times$45\\
			Deconv$_{32}$ & Deconv, BN, PReLU & 256$\times$55$\times$45\\
			\hline
			UpSample$_{2}$ & UpSampling & 256$\times$109$\times$89\\
			Deconv$_{21}$ & Deconv, BN, PReLU & 128$\times$109$\times$89\\
			Deconv$_{22}$ & Deconv, BN, PReLU & 128$\times$109$\times$89\\
			\hline
			UpSample$_{1}$ & UpSampling & 128$\times$218$\times$178\\
			Deconv$_{11}$ & Deconv, BN, PReLU & 64$\times$218$\times$178\\
			Deconv$_{12}$ & Deconv, BN, PReLU & 64$\times$218$\times$178\\
			Deconv$_{13}$ & Deconv, BN, PReLU & 7$\times$218$\times$178\\		
			\hline
	\end{tabular}}
	\caption{Configuration of the Semantic Segmentation Network. For all the convolution/ deconvolution layers, kernel size and stride values are respectively set to 3 and 1. To prevent confusion, we are not showing the side loss layers, namely Deconv$_{43}$, Deconv$_{33}$ and Deconv$_{23}$.}\label{tab:segnetconfig}
	\vspace{-4mm}
\end{table}

\section{Experimental Results}\label{sec:experimentalresults}

\subsection{Training Semantic Segmentation Network}\label{sec:trainsemanticseg}
\begin{figure}
	\centering
	\includegraphics[width=0.47\textwidth]{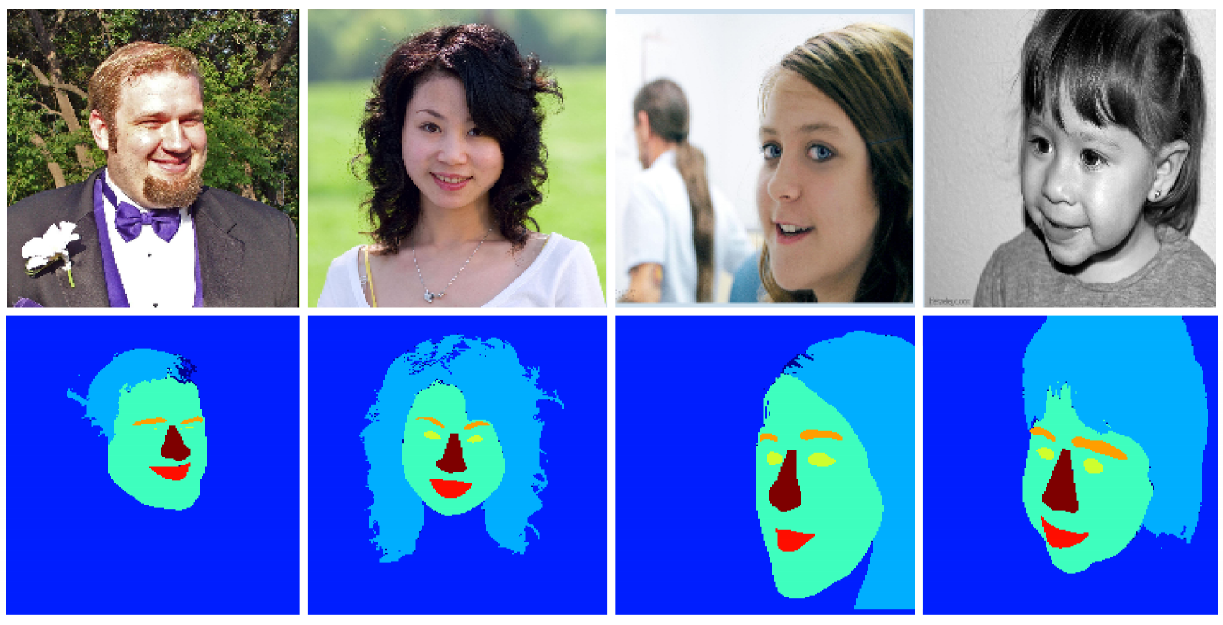}
	\caption{Examples of the Helen face dataset \cite{le2012interactive} supplemented with segment label annotations \cite{smith2013exemplar} and then grouped into 7 semantic classes. In bottom row, colors indicate different class labels.} \label{fig:helenexample}
	\vspace{-4mm}
\end{figure}

\begin{figure}
	\centering
	\includegraphics[width=0.47\textwidth]{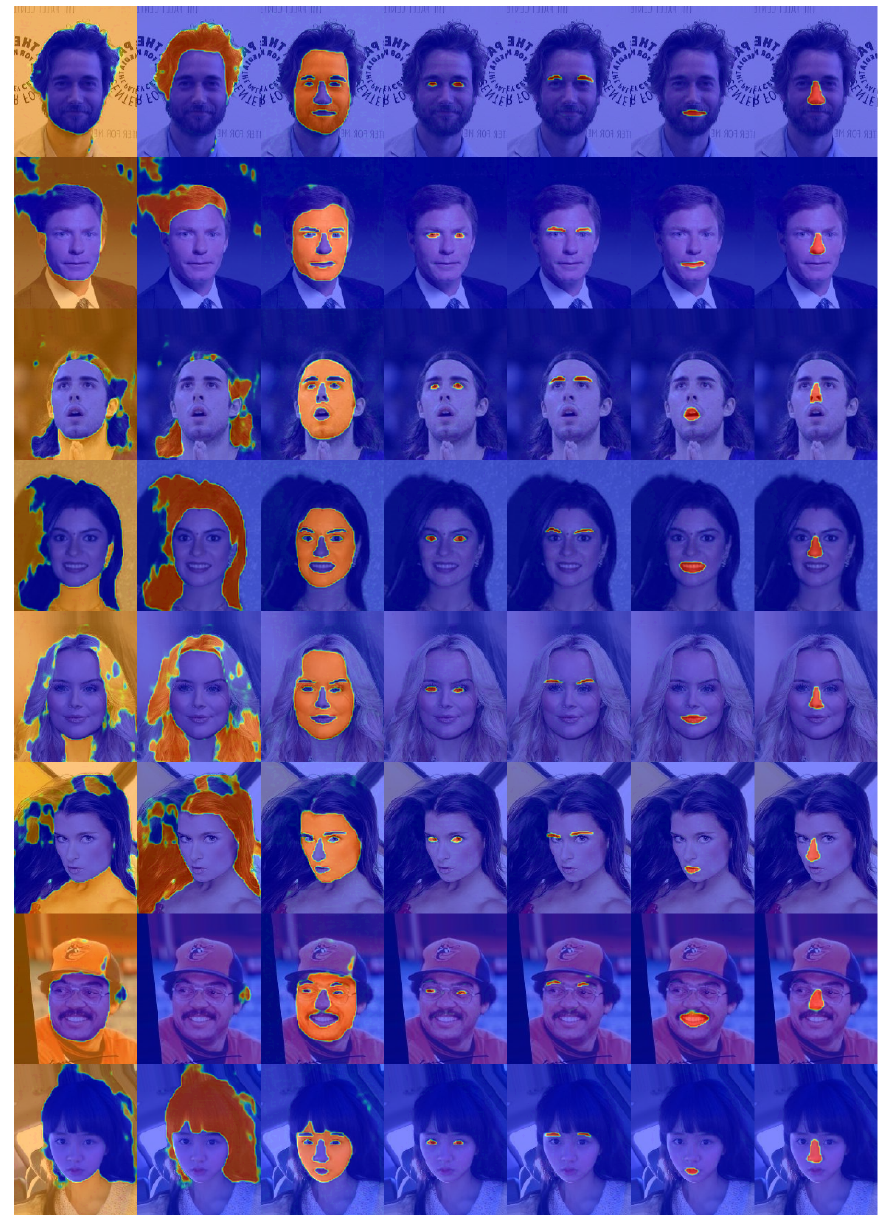}
	\caption{Examples of the segmentation masks generated by our semantic segmentation network for previously unseen images. From left to right: background, hair, face skin, eyes, eyebrows, mouth and nose.} \label{fig:segnetexample}
	\vspace{-4mm}
\end{figure}

In this paper, we are interested in facial attribute prediction. Hence, face parsing problem \cite{smith2013exemplar} which aims at pixel-level classification of a face image into multiple semantic regions is a suitable auxiliary task for us. To train the semantic segmentation network, we begin with 11 segment label annotations per image that \cite{smith2013exemplar} provides to supplement Helen face dataset \cite{le2012interactive}. These labels are as follows: background, face skin (excluding ears and neck), left eyebrow, right eyebrow, left eye, right eye, nose, upper lip, inner mouth, lower lip and hair. We combine left and right eye (eyebrow) labels to create a single eye (eyebrow) label. Similarly, we aggregate upper lip, inner mouth, and lower lip to generate a single mouth label. As a result we end up with a total of 7 labels (background, hair, face skin, eyes, eyebrows, mouth and nose). Figure \ref{fig:helenexample} illustrates a few instances of the input images along with their corresponding segment label annotations. The face parsing dataset \cite{smith2013exemplar} comes with 2,330 images in three splits of 2000, 230 and 100, respectively for training, validation and test. However, for the attribute prediction task, we can use the entire dataset to train the semantic segmentation network. We train our model with softmax cross entropy loss. Section \ref{sec:implementationdetails} provides details on the training procedure. Figure \ref{fig:segnetexample} shows a few examples of segmentation maps generated by our network. Despite very few number of training data used in its training process, the semantic segmentation network is able to successfully localize various facial regions in previously unseen images. Later, we evaluate our proposed attribute prediction model where these semantic segmentation cues are utilized to improve facial attribute prediction.

\subsection{Datasets and Evaluation Metrics}
We mainly evaluate our proposed approach on the CelebA dataset \cite{liu2015faceattributes}. CelebA consists of 202,599 images partitioned into training, validation and test splits with approximately 162K, 20K and 20K images in the respective splits. There are a total of 10K identities (20 images per identity) with no identity overlap between evaluation splits. Images are annotated with 40 facial attributes such as, ``wavy hair", ``mouth slightly open", ``big lips", etc. In addition to the original images, CelebA provides a set of pre-cropped images. We report our results on both of these image sets. It is worth noting that Liu \textit{et al.} \cite{liu2015faceattributes} have used both the training and validation data in order to train different parts of their model. In particular, training data has been used to pre-train and fine-tune ANet and LNet while they train SVM classifiers using the validation data. In our experiments, we only use the training split to train our attribute prediction networks.

To supplement the analyses on CelebA dataset \cite{liu2015faceattributes}, we also provide experimental results on LFWA\cite{liu2015faceattributes}. LFWA has a total of 13,232 images of 5,749 identities with pre-defined train and test splits which divide the entire dataset into two approximately equal partitions. Each image is annotated with the same 40 attributes used in CelebA\cite{liu2015faceattributes} dataset. For the LFWA dataset \cite{liu2015faceattributes}, we follow the same evaluation protocol as the one for CelebA dataset \cite{liu2015faceattributes}. 

To evaluate the attribute prediction performance, Liu \textit{et al.} \cite{liu2015faceattributes} use classification accuracy/error. However, we believe that due to significant imbalance between the numbers of positive and negatives instances per attribute, such measure cannot appropriately evaluate the quality of different methods. Similar point has been raised by \cite{rudd2016moon, huang2016learning} as well. Therefore, in addition to the classification error, we also report the average precision of the prediction scores.

\subsection{Evaluation of Facial Attribute Prediction}
For all the numbers reported here, we want to point out that FaceTracer \cite{kumar2008facetracer} and PANDA \cite{zhang2014panda} use groundtruth landmark points to attain face parts. Wang \textit{et al.} \cite{wang2016walk} use 5 million auxiliary image pairs, collected by the authors, to pre-train their model. Wang \textit{et al.} \cite{wang2016walk} also use state-of-the-art face detection and alignment to extract the face region from CelebA and LFWA images. However, we train all our models from scratch with only attribute labels and the auxiliary face parsing labels.

\subsubsection{Evaluation on CelebA dataset}
We compare our proposed method with the existing state-of-the-art attribute prediction techniques on the CelebA dataset \cite{liu2015faceattributes}. To prevent any confusion and have a fair comparison, Table \ref{tab:CelebAresult} reports the performances in two separate columns distinguishing the experiments that are conducted on the original image set from those where the pre-cropped image set have been used. We see that even our basic model with global average pooling, with the exception of the MOON \cite{rudd2016moon}, outperforms previous state-of-the-art techniques. Accordingly, we can make two observations.

First, a simple yet well designed architecture can be very effective. Liu \textit{et al.} \cite{liu2015faceattributes} combine three deep convolutional neural networks with SVM and Rudd \textit{et al.} \cite{rudd2016moon} have adopted VGG16 \cite{simonyan2014very} topped with a novel objective function. These models are drastically larger than our basic network. Specifically, in \cite{liu2015faceattributes}, LNet$_o$ and LNet$_s$ have network structures similar to AlexNet \cite{krizhevsky2012imagenet}. AlexNet has 60M parameters. Thus, only the localization part in \cite{liu2015faceattributes}, not considering ANet, has a total of 120M parameters. Rudd \textit{et al.} \cite{rudd2016moon} adopt VGG16 \cite{simonyan2014very} that has 138M parameters. Our basic attribute prediction network has only 24M parameters thanks to replacing fully connected layers with a single global average pooling.

Second, \cite{rudd2016moon} and \cite{liu2015faceattributes} are built on the top of networks previously trained on massive object category (and facial identity) data while we train all our networks from scratch. Hence, we reject the necessity of pre-training on other large scale benchmarks, arguing that CelebA dataset \cite{liu2015faceattributes} itself is sufficiently large for successfully training facial attribute prediction models from scratch.

Experimental results indicate that under different settings and evaluation protocols, our proposed semantic segmentation-based pooling and gating mechanisms can be effectively used to boost the facial attribute prediction performance. That is particularly important given that our global average pooling baseline already beats the majority of the existing state-of-the-art methods. To see if SSP and SSG are complementary to each other, we also report their combination where the corresponding predictions are simply averaged. We observe that such process further boosts the performance. 

\begin{table}
	\centering
	\begin{tabular}{lcc}
		\toprule
		\multicolumn{3}{c}{\textbf{Classification Error\%}}\\
		\midrule
		Method & Original & Pre-cropped\\
		\midrule
		FaceTracer \cite{kumar2008facetracer} & 18.88 & --\\
		PANDA \cite{zhang2014panda} & 15.00 & --\\		
		Liu \textit{et al.} \cite{liu2015faceattributes} & 12.70 & --\\
		Wang \textit{et al.} \cite{wang2016walk} & 12.00 & --\\
		Zhong \textit{et al.} \cite{zhong2016leveraging} & 10.20 & --\\
		Rudd \textit{et al.} \cite{rudd2016moon}: Separate & -- & 9.78\\
		Rudd \textit{et al.} \cite{rudd2016moon}: MOON & -- & 9.06\\
		SPPNet$^{*}$ & -- & 9.49\\		
		Naive Approach & 9.62 & 9.13\\
		BBox & -- & 8.76\\
		Ours: Avg. Pooling & 9.83 & 9.14\\
		Ours: SSG &  9.13 & 8.38\\
		Ours: SSP & 8.98 & 8.33 \\
		Ours: SSP + SSG &  8.84 & 8.20\\
		\toprule
		\multicolumn{3}{c}{\textbf{Average Precision\%}}\\
		\midrule
		Method & Original & Pre-cropped\\
		\midrule
		SPPNet$^{*}$ & -- & 77.69\\
		Naive Approach & 76.29 & 79.74\\
		BBox & -- & 79.95\\
		Ours: Avg. Pooling & 77.16 & 79.74\\
		Ours: SSG & 77.46 & 80.55\\		
		Ours: SSP & 78.01 & 81.02\\
		Ours: SSP + SSG & 78.74 & 81.45\\
		\toprule
		\multicolumn{3}{c}{\textbf{Balanced Accuracy\% \cite{huang2016learning}}}\\
		\midrule
		Method & Original & Pre-cropped\\
		\midrule
		Huang \textit{et al.} \cite{huang2016learning} & -- & 84.00\\
		Ours: Avg. Pooling & -- & 86.73\\
		Ours: SSG & -- & 87.82\\
		Ours: SSP & -- & 88.24\\					
		\bottomrule\\
	\end{tabular}
	\caption{Attribute prediction performance evaluated by the classification error, average precision and balanced classification accuracy \cite{huang2016learning} on the CelebA \cite{liu2015faceattributes} original and pre-cropped image sets.}\label{tab:CelebAresult}
	\vspace{-4mm}
\end{table}

To investigate the importance of aggregating features within the semantic regions, we replace the global average pooling in our basic model with the spatial pyramid pooling layer \cite{he2014spatial}. We use a pyramid of two levels and refer to this baseline as SPPNet$^*$. While aggregating the output activations in different locations, SPPNet$^*$ does not align its pooling regions according to the semantic context that appears in the image. This is in direct contrast with the intuition behind our proposed methods. Experimental results shown in Table \ref{tab:CelebAresult} confirm that simply pooling the output activations at multiple locations is not sufficient. In fact, it results in a lower performance than global average pooling. This verifies that the improvement obtained by our proposed models is due to their content aware pooling/gating mechanisms.

\textbf{Naive Approach }A naive alternative approach is to consider the segmentation maps as additional input channels. To evaluate its effectiveness, we feed the average pooling basic model with 10 input channels, 3 for RGB colors and 7 for different semantic segmentation maps. The input is normalized using Batch Normalization \cite{ioffe2015batch}. We train the network using the same setting as other aforementioned models. Our experimental results indicate that such naive approach cannot leverage the localization cues as good as our proposed methods. Table \ref{tab:CelebAresult} shows that at best, the naive approach is on par with the average pooling basic model. We emphasize that feeding semantic segmentation maps along with RGB color channels to a convolutional network results in blending the two modalities in an \textit{addition} fashion. Instead, our proposed mechanisms take a \textit{multiplication} approach by masking the activations using the semantic regions.

\textbf{Semantic Masks vs. Bounding Boxes }To analyze the necessity of semantic segmentation, we generate a baseline, namely BBox, which is similar to SSP. However, we replace the semantic regions in SSP with the bounding boxes on the facial landmarks. Note that we use the groundtruth location of the facial landmarks, provided in CelebA dataset \cite{liu2015faceattributes}, to construct the bounding boxes. Hence, to some extent, the performance of BBox is the upper bound of the bounding box experiment. There are 5 facial landmarks including left eye, right eye, nose, left mouth and right mouth. We use boxes with area $20^2$ ($40^2$ gives similar results) and 1:1, 1:2 and 2:1 aspect ratios. Thus, there are a total of 16 regions including the whole image itself. From Table \ref{tab:CelebAresult}, we see that our proposed models, regardless of the evaluation measure, outperform the bounding box alternative suggesting that semantic masks should be favored over the bounding boxes on the facial landmarks.

\textbf{Balanced Classification Accuracy } Given the significant imbalance in the attribute classes, also noted by \cite{huang2016learning, rudd2016moon}, we suggested using average precision instead of classification accuracy/error to evaluate attribute prediction. Instead, Huang \textit{et al.} \cite{huang2016learning} have adopted balanced accuracy measure. To see if our proposed approach is superior to \cite{huang2016learning} under balanced accuracy measure, we fine-tuned our models with the weighted ($\propto$ imbalance level) binary cross entropy loss. From Table \ref{tab:CelebAresult}, we observe that under balanced accuracy \cite{huang2016learning}, all the variations of our proposed model outperform \cite{huang2016learning} with large margins.

\begin{table}
	\centering
	\begin{tabular}{lcc}
		\toprule
		Method & Classification Error\%& AP\%\\
		\midrule
		FaceTracer \cite{kumar2008facetracer} & 26.00 & --\\
		PANDA \cite{zhang2014panda} & 19.00 & --\\		
		Liu \textit{et al.} \cite{liu2015faceattributes} &  16.00 & --\\
		Zhong \textit{et al.} \cite{zhong2016leveraging} & 14.10 & --\\
		Wang \textit{et al.} \cite{wang2016walk} & 13.00 & --\\
		Ours: Avg. Pooling & 14.73 & 82.69\\
		Ours: SSG &  13.87 & 83.49\\
		Ours: SSP & 13.20 & 84.53\\
		Ours: SSP + SSG &  12.87 & 85.28\\
		\bottomrule\\
	\end{tabular}
	\caption{Attribute prediction performance evaluated by the classification error and the average precision (AP) on LFWA \cite{liu2015faceattributes} dataset.}\label{tab:resultsLFWA}
\end{table}

\subsubsection{Evaluation on LFWA dataset} 
To better understand the effectiveness of our proposed approach, we report experimental results on the LFWA dataset \cite{liu2015faceattributes} in Table \ref{tab:resultsLFWA}. We observe that, all the models proposed in this work which exploit localization cues improve our basic model. Specifically, SSP + SSG achieves considerably better performance than the average pooling basic model with 1.86\% in classification error and 2.59\% in the average precision. Our best model also outperforms all other state-of-the-art methods.

\begin{table}
	\centering
	\begin{tabular}{lcc}
		\toprule
		Region & w/o Attributes & w/ Attributes\\
		\midrule
		Bakcground & 89.25 & 89.64\\
		Hair & 47.56 & 48.32\\
		Face skin & 78.65 & 79.92\\
		Eyes & 46.83 & 56.33\\
		Eyebrows & 31.22 & 42.25\\
		Mouth & 62.03 & 65.42\\
		Nose & 77.40 & 77.74\\
		\textbf{Average} & 61.84 & 65.66\\
		\bottomrule\\
	\end{tabular}
	\caption{Effect of facial attributes on semantic face parsing performance evaluated by Intersection over Union (IoU\%).}\label{tab:faceparsing}
\end{table}

\subsection{Facial Attributes for Semantic Face Parsing}
In this work, we established how semantic segmentation can be used to improve facial attribute prediction. What if we reverse the roles. Can facial attributes improve semantic face parsing? To evaluate this, we jointly train two networks where the first 8 layers of our basic attribute prediction network share weights with the encoder part of the semantic segmentation network. We optimize w.r.t the aggregation of two losses. Specifically, the attribute prediction loss on the CelebA \cite{liu2015faceattributes} dataset and the semantic segmentation loss on the Helen face \cite{le2012interactive} dataset using facial segment labels of \cite{smith2013exemplar}. We follow pre-defined data partitions of \cite{smith2013exemplar}, detailed in section \ref{sec:trainsemanticseg}, and use Intersection over Union (IoU) as the evaluation measure. Table \ref{tab:faceparsing} shows nearly 4\% boost when attributes are incorporated, indicating the positive effect of attributes in improving semantic face parsing. This shows that there exist an interrelatedness between attribute prediction and semantic segmentation. In future, we will further explore this promising direction.

\begin{figure}
	\centering
	\includegraphics[width=0.48\textwidth]{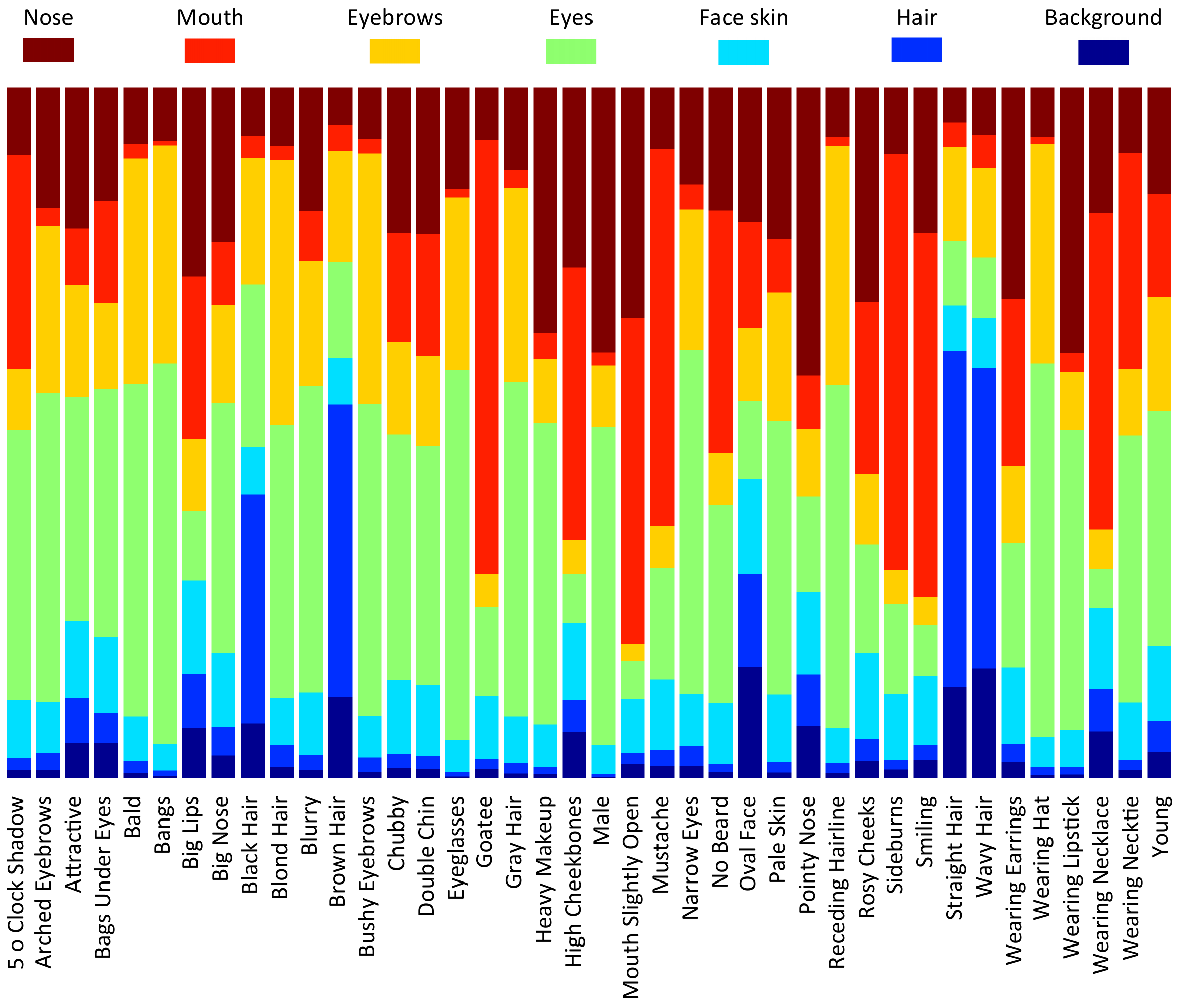}
	\caption{Contribution of different semantic regions in attribute prediction as learned by the localization branch of SSP. Values are averaged over multiple random mini-batches of 32 images.} \label{fig:localizationweights}
	\vspace{-6mm}
\end{figure}

\begin{figure*}
	\centering
	\includegraphics[width=1\textwidth]{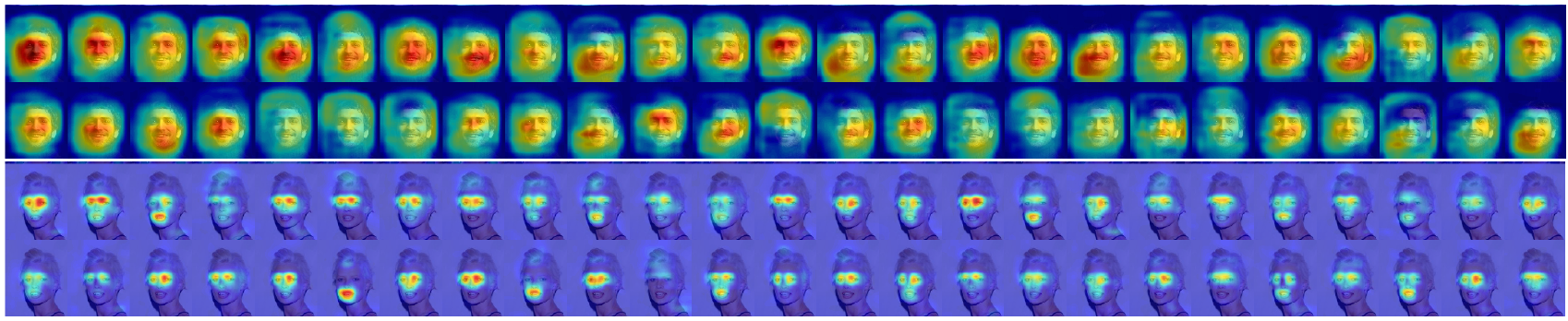}
	\caption{Top fifty activation maps of the last convolution layer sorted in descending order w.r.t the average activation values. Top: Global average pooling. Bottom: SSP.}\label{fig:outputactivation}
	\vspace{-6mm}
\end{figure*}

\subsection{Visualizations}
Figure \ref{fig:localizationweights} illustrates per-attribute weights that the localization branch of the SSP has learned in order to combine the predictions associated with different semantic regions. We observe that attributes such as ``Black Hair", ``Brown Hair", ``Straight Hair" and ``Wavy Hair" have strong bias towards the hair region. This matches our expectation. However, attribute ``Blond Hair" does not behave similarly. We suspect that it is because the semantic segmentation network does not perform as consistent on light hair colors as it does on the dark ones (refer to Figure \ref{fig:segnetexample}). Attributes such as ``Goatee", ``Mouth Slightly Open", ``Mustache" and ``Smiling" are also showing a large bias towards the mouth region. While these are aligned with our human knowledge, ``Sideburns" and ``Wearing Necklace" apparently have incorrect biases. Unlike the global pooling which equally affects a rather large spatial domain, we expect SSP to generate activations that are semantically aligned. To evaluate our hypothesis, in Figure \ref{fig:outputactivation}, we show the activations for the top fifty channels of the last convolution layer. Top row corresponds to our basic network with global average pooling while the bottom row is generated when we replace global average pooling with SSP. We observe that, activations generated by SSP are clearly more localized than those obtained from the global average pooling.

\section{Implementation Details}\label{sec:implementationdetails}
All of our experiments were conducted on a single NVIDIA Titan X GPU.
We use AdaGrad \cite{duchi2011adaptive} with mini-batches of size 32 to train the attribute prediction models from scratch. The learning rate and weight decay are respectively set to 0.001 and 0.0005. We follow the same setting for training the semantic segmentation network. We perform data augmentation by randomly flipping (horizontally) the input images. In SSP experiments, we resize the output of the semantic segmentation network at Deconv$_{23}$ layer to 14$\times$12 (resolution of the final convolution layer). To do so, we use max and average pooling operations. Since max pooling increases the spatial support of the region, we use it for the masks associated with eyes, eyebrows, nose and mouth. This helps us to capture some context as well. We use average pooling for the remaining regions. For SSG experiments, we use the output of Deconv$_{33}$ layer, in the semantic segmentation network, as the localization cue. The attribute prediction and semantic segmentation networks are respectively trained for 40K and 75K iterations.

\section{Conclusion}\label{sec:conclusion}
Aligned with the trend of part-based attribute prediction methods, we proposed employing semantic segmentation to improve facial attribute prediction. Specifically, we transfer localization cues from the auxiliary task of semantic face parsing to the facial attribute prediction problem. In order to guide the attention of our attribute prediction model to the regions which different attributes naturally show up, we introduced SSP and SSG. While SSP is used to restrict the aggregation procedure of final activation maps to regions that are semantically consistent, SSG carries the same notion but applies it to the earlier layers. We evaluated our proposed methods on CelebA and LFWA datasets and achieved state-of-the-art performance. We also showed that facial attributes can improve semantic face parsing. We hope that this work encourages future research efforts to invest more in the interrelatedness of these two problems.\\

\noindent\textbf{Acknowledgments: }\small{We thank anonymous reviewers for insightful feedback, and Amir Emad, Shervin Ardeshir and Shayan Modiri Assari for fruitful discussions. Mahdi M. Kalayeh and Mubarak Shah are partially supported by NIJ W911NF-14-1-0294. Boqing Gong is supported in part by NSF IIS \#1566511 and thanks Adobe Systems for a gift.}

{
\bibliographystyle{ieee}
\bibliography{camera_ready.bib}
}

\end{document}